\documentclass[conference]{IEEEtran}
\IEEEoverridecommandlockouts
\usepackage{amsmath}
\usepackage{subcaption}
\usepackage{cite}
\usepackage{amsmath,amssymb,amsfonts}
\usepackage{algorithmic}
\usepackage{graphicx}
\usepackage{textcomp}
\usepackage{url}
\usepackage{xcolor}
\usepackage{lscape} 
\usepackage{multirow} 
\def\BibTeX{{\rm B\kern-.05em{\sc i\kern-.025em b}\kern-.08em
    T\kern-.1667em\lower.7ex\hbox{E}\kern-.125emX}}

\begin{document}
\title{OptiGait-LGBM: An Efficient Approach of Gait-based Person Re-identification in Non-Overlapping Regions}

\author{\IEEEauthorblockN{Md. Sakib Hassan Chowdhury}
\IEEEauthorblockA{\textit{Mechatronics Engineering} \\
\textit{Rajshahi University of Engineering \& Technology}\\
Rajshahi, Bangladesh \\
sakib.mte.ruet@gmail.com}
\and
\IEEEauthorblockN{Md. Hafiz Ahamed}
\IEEEauthorblockA{\textit{Mechatronics Engineering} \\
\textit{Rajshahi University of Engineering \& Technology}\\
Rajshahi, Bangladesh \\
hafiz@mte.ruet.ac.bd}
\and
\IEEEauthorblockN{Bishowjit Paul}
\IEEEauthorblockA{\textit{Mechatronics Engineering} \\
\textit{Rajshahi University of Engineering \& Technology}\\
Rajshahi, Bangladesh\\
bishowjitpaul6@gmail.com}
\and
\IEEEauthorblockN{Sarafat Hussain Abhi}
\IEEEauthorblockA{\textit{Mechatronics Engineering} \\
\textit{Rajshahi University of Engineering \& Technology}\\
Rajshahi, Bangladesh \\
abhi@mte.ruet.ac.bd}
\and

\IEEEauthorblockN{Abu Bakar Siddique}
\IEEEauthorblockA{\textit{Mechatronics Engineering} \\
\textit{Rajshahi University of Engineering \& Technology}\\
Rajshahi, Bangladesh \\
1808031@student.ruet.ac.bd}
\and
\IEEEauthorblockN{Md. Robius Sany}
\IEEEauthorblockA{\textit{Mechatronics Engineering} \\
\textit{Rajshahi University of Engineering \& Technology}\\
Rajshahi, Bangladesh \\
jitpaul.khulna@gmail.com}
}

\maketitle

\begin{abstract}
Gait recognition, known for its ability to identify individuals from a distance, has gained significant attention in recent times due to its non-intrusive verification. In terms of video gait-based identification systems excel on large public datasets, their performance drops with real-world unconstrained gait data due to various factors. Among them, uncontrolled outdoor non-overlapping situations with different illuminations and computational efficiency are one of the core problems in terms of gait based authentication. We found no such datasets to address all these challenges all at a time. In this paper, we propose an OptiGait-LGBM model capable of recognizing person's re-identification under these challenges using skeletal model approach which has one of the prime benefit like person's appearance inconsistency by product. The model constructs a dataset from landmark positions, minimizing memory usage with non-sequential data. A bechmark dataset, RUET-GAIT is introduced while it addresses uncontrolled gait sequence in outdoor complex environment. The process involves extracting skeletal joint landmarks, generating a numerical datasets, and developing an OptiGait-LGBM gait classification model. Our challenge is to address all this complex situatuion addressed earlier with minimal computational cost from its own kind. A comparative analysis with these ensemble techniques with Random Forest and CatBoost demonstrates the out-performance of the proposed approach in terms of accuracy as well as memory usage and training time. The proposed method provides a novel, low-cost, memory-efficient video-based recognition solution for real-world scenarios.
\end{abstract}


\section{Introduction}

Person identification has been pivotal over the past two decades, playing a crucial role in applications such as surveillance, forensics, and access control. Among the three primary types of individual authentication—knowledge-based, object-based, and biometric—biometric systems have gained prominence due to their reliance on physiological and behavioral traits, including fingerprints, iris scans, speech patterns, and gait patterns. Notably, gait recognition, which identifies individuals based on their unique walking style, offers distinct advantages due to its remote operability and robustness in real-world scenarios.

However, despite these benefits, gait recognition faces several challenges, including variations in walking surfaces \cite{connor2018biometric}, speed \cite{hortobagyi2015effects} , clothing \cite{yeoh2016clothing}, injuries \cite{tao2012gait}, and viewing angles \cite{wan2018survey}. To address these issues, recent research has focused on developing robust gait representations that can effectively adapt to these variables. This focus is essential, as biometric-based identification systems work by capturing samples, extracting relevant features, and comparing them to registered features.

A critical aspect of intelligent surveillance is person re-identification (Re-ID), which involves matching images of the same individual captured at different times and locations. This process ensures continuous tracking, even in the presence of trajectory discontinuities. Unlike traditional recognition, which retrieves known identities from a database, Re-ID matches individuals without requiring prior knowledge of their identities. This can be done in a supervised learning framework, where labeled data is used for training. Such capabilities are particularly useful for wide-area surveillance applications, further emphasizing the importance of effective gait recognition systems. Central to the effectiveness of these systems is gait analysis, which captures detailed walking patterns \cite{boyd2005biometric}. Researchers extract vital metrics such as step length, stride length, step width, and foot angle to gain insight into human locomotion. This information not only helps improve biometric systems, but also has significant implications in various fields, including clinical diagnoses, rehabilitation, biomechanics, sports science, ergonomics, and forensics \cite{benabdelkader2002person}. By bridging these areas, advances in gait recognition contribute to a deeper understanding of both individual identification and human movement dynamics.

The phrase "Gait Cycle" (shown in Figure 1) is often used to describe the process of capturing a human walk. Since walking entails a repetitive pattern of varying the angular \cite{zhang2004human} and linear dimensions of the body, various sections of the body must be used at different times.

\begin{figure}[h] \centering \includegraphics[width=0.56\textwidth]{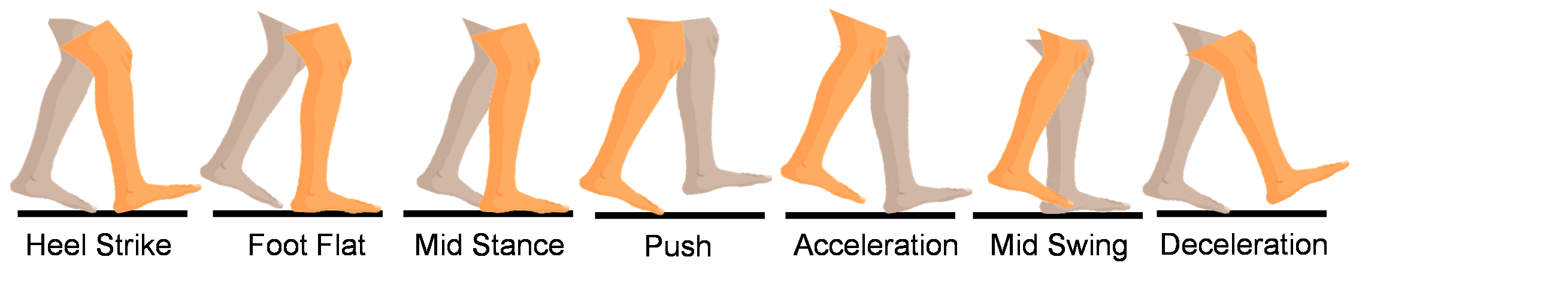} \caption{Human Gait Cycle} \end{figure}

Furthermore, gait data acquisition can be vision-based or sensor-based. Vision-based systems utilize cameras for comprehensive data on posture and gait dynamics \cite{gaya2024deep}, while sensor-based systems employ wearable sensors for real-time collection \cite{shi2024review}. Both methods offer unique advantages for continuous monitoring and precise analysis, with vision-based methods including marker-based and marker-free systems \cite{zhang2024unmanned}, and sensor-based methods using floor sensors and wearable sensors for flexibility in research and clinical applications \cite{dammeyer2024classification}. 

In terms of configuration, vision-based data acquisition systems can be either overlapping or non-overlapping. Overlapping systems, common in video surveillance, use multiple cameras with intersecting fields of view to provide continuous coverage and reduce blind spots. In contrast, non-overlapping systems, typical in real-life scenarios, present challenges for gait recognition due to gaps in coverage and the need to match data from different cameras. To overcome these challenges, advanced algorithms, data fusion techniques, and careful system design are necessary to ensure accurate and reliable identification. Figure 2 illustrates the difference between overlapping and non-overlapping camera configurations, highlighting their distinct capabilities in capturing visual information.

\begin{figure}[h] \centering \includegraphics[width=0.5\textwidth]{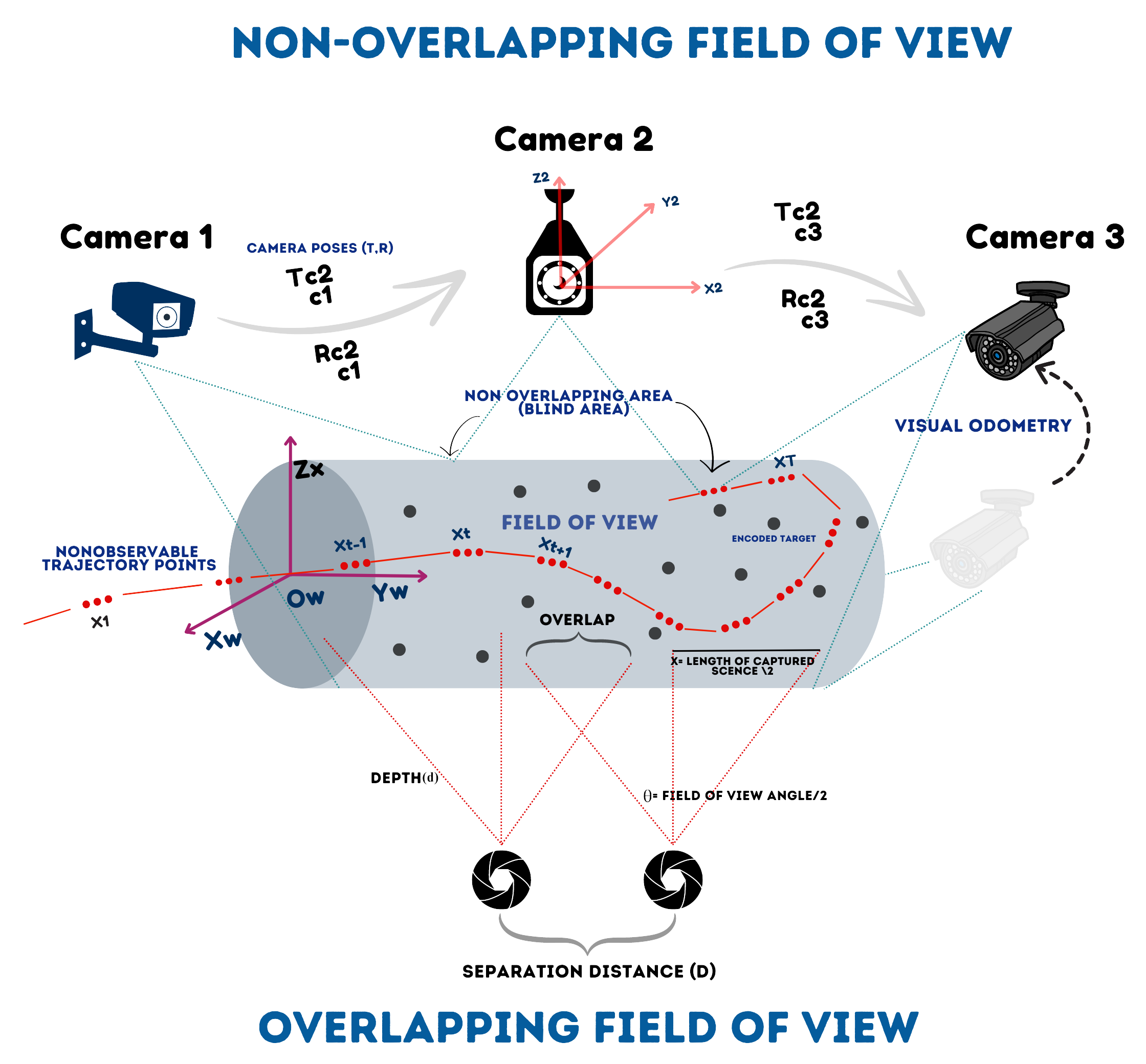} \caption{Overlapping vs Non-Overlapping Camera} \end{figure}

Gait analysis supports various fields, primarily healthcare, by aiding in diagnosing and rehabilitating mobility issues, leading to personalized treatments. In biomechanics, it helps develop advanced prosthetics and sports equipment by understanding walking and running mechanics \cite{khatavkar2022gait}. Sports scientists utilize gait analysis to improve athletic performance and reduce injuries \cite{melo2020does}. Moreover, in ergonomics, it enhances workplace design by evaluating movement, reducing strain, and increasing productivity \cite{genitrini2022impact}. The security industry employs gait analysis for biometric authentication, strengthening access control and cybersecurity \cite{narayan2023deep}. Exoskeleton and robotics research relies on it to mimic human gait, which is essential for developing walking robots and exoskeletons \cite{zhong2020synchronous}. Aging research uses gait analysis to preserve mobility in seniors, while forensic science employs it for identifying individuals in criminal investigations \cite{kahya2019brain, macoveciuc2019forensic}. Thus, gait identification technologies—computer vision-based \cite{ng2020measuring}, wearable sensor-based \cite{prasanth2021wearable}, and floor sensor-based systems \cite{alharthi2021spatiotemporal}—each offer unique benefits, making gait analysis crucial across multiple domains.

In this context, our research addresses the need for an effective biometric verification system based on distinctive gait patterns, particularly in uncontrolled environments with irregular data. The primary objectives are to develop a robust system with low computational and memory costs and to create a model resilient to variations in viewpoint and walking path.

To achieve this, the outline includes an introduction to biometric systems and data acquisition, a review of literature on gait re-identification methodologies, details on dataset acquisition, a comprehensive methodology for the proposed gait re-identification method, an analysis of results using various models, and a discussion of limitations and future scopes.

\section{Literature Review}
The literature on gait recognition explores various methods for person re-identification, focusing on both vision-based and sensor-based techniques. Early studies, starting in the 1970s with Johansson et al. \cite{johansson1973visual}, demonstrated the potential of gait as a biometric identifier, paving the way for further exploration in this field. The HumanID project \cite{banakou2007human} and advancements in time-of-flight cameras \cite{zhang2014gait} significantly enhanced gait analysis, leading to high recognition accuracies in recent studies \cite{wan2017gait, wang2020gait}.

Gait recognition methods can be broadly categorized into model-based approaches, which utilize detailed mathematical models, and model-free techniques that rely on general gait characteristics. Model-based methods, such as those developed by Hamdoun and Chabchoub \cite{hamdoun2015gait}, offer high accuracy but often require complex models that can be difficult to implement in real-world scenarios. Conversely, model-free techniques, which utilize silhouette-based features, provide greater flexibility and ease of use \cite{benabdelkader2002person, yang2019survey}.

Recent advancements have introduced deep learning approaches to gait recognition, leveraging convolutional neural networks (CNNs) and recurrent neural networks (RNNs) for feature extraction and classification. These methods have shown promising results, outperforming traditional approaches in terms of accuracy and robustness \cite{li2019review, zhou2021deep}. For example, Liu et al. \cite{liu2019gait} demonstrated that deep learning models could effectively capture the temporal dynamics of gait, leading to improved performance under varying conditions.

Moreover, the integration of multiple modalities, such as combining gait with other biometric features like face recognition, has been explored to enhance re-identification accuracy. Studies by Liu et al. \cite{liu2020gait} highlight the advantages of multi-modal systems in improving robustness against variations in environmental conditions and occlusions.

The impact of environmental factors on gait recognition has also been a focus of recent research. Studies by Khan et al. \cite{khan2021gait} emphasized the challenges posed by variations in walking surfaces, speeds, and clothing. To address these issues, methods such as data augmentation and domain adaptation are being explored to enhance model robustness \cite{zhang2020gait}. Furthermore, recent works emphasize the importance of feature selection and dimensionality reduction techniques to improve computational efficiency without sacrificing recognition accuracy \cite{wu2021gait}.

The growing interest in gait recognition is reflected in its diverse applications, including security and surveillance, healthcare, and human-computer interaction. Gait analysis plays a crucial role in healthcare, aiding in the diagnosis and rehabilitation of mobility issues \cite{khatavkar2022gait}. Additionally, the field of sports science employs gait recognition to enhance athletic performance and prevent injuries \cite{melo2020does}. In forensic science, gait patterns are used to identify individuals in criminal investigations, underscoring the importance of this research \cite{kahya2019brain}.

Recent studies have also begun to explore the ethical implications and privacy concerns associated with gait recognition technologies. Research by Prabhakar et al. \cite{prabhakar2020gait} raises important questions regarding consent and data protection, emphasizing the need for responsible use of biometric systems.

Overall, the field of gait recognition is evolving rapidly, with ongoing research focused on developing robust systems that adapt to real-world variations in gait. By combining insights from various methodologies and leveraging advanced computational techniques, researchers aim to enhance the accuracy and applicability of gait recognition systems across multiple domains. 

In this work, we make several key contributions to advance machine learning applications- ensemble learning techniques. First, we prioritize traditional machine learning (ML) techniques over deep learning (DL), enabling reduced memory and data requirements for training and testing, which leads to significantly shorter training times. We introduce a customized dataset tailored to our specific application and propose a novel ML model that effectively utilizes this data. Additionally, we incorporate a camera correlation factor to enhance accuracy across different perspectives and leverage multi-view data integration to create a coherent representation in non-overlapping situations. Our model is designed to perform robustly in complex backgrounds and varying lighting conditions, addressing challenges often overlooked in existing datasets. Through these contributions, we aim to provide an efficient and effective solution for real-world scenarios.

\section{Dataset Acquisition}
To effectively re-identify gaits in real-life situations, we created a dataset capturing gait patterns in uncontrolled environments. This section outlines our data collection process and equipment, emphasizing why our dataset offers advantages over the CASIA-B dataset \cite{iaCenterBiometrics}.

\subsection{Surveillance Camera Setup \& Footage Acquisition}\label{AA}
Accurate gait pattern recognition requires well-positioned surveillance cameras to capture distinct side-view gait signatures. We used a mobile camera setup at Uporvadra, Kazla, Rajshahi-6204, Bangladesh. The setup included:

\begin{itemize}
    \item \textbf{Camera 1 (Entry Point Observer):} Captures the initial gait patterns of individuals as they enter the monitored area.
    \item \textbf{Camera 2 (Transition Observer):} Records gait patterns during the mid-transition phase through the surveillance zone.
    \item \textbf{Camera 3 (Central Observer):} Provides comprehensive coverage of the gait cycle within the central area of the monitored zone.
    \item \textbf{Camera 4 (Exit Point Observer):} Monitors the final stages of gait as individuals exit the surveillance area for accurate re-identification.
\end{itemize}

\begin{figure}[h]
\centering
\includegraphics[width=0.3\textwidth]{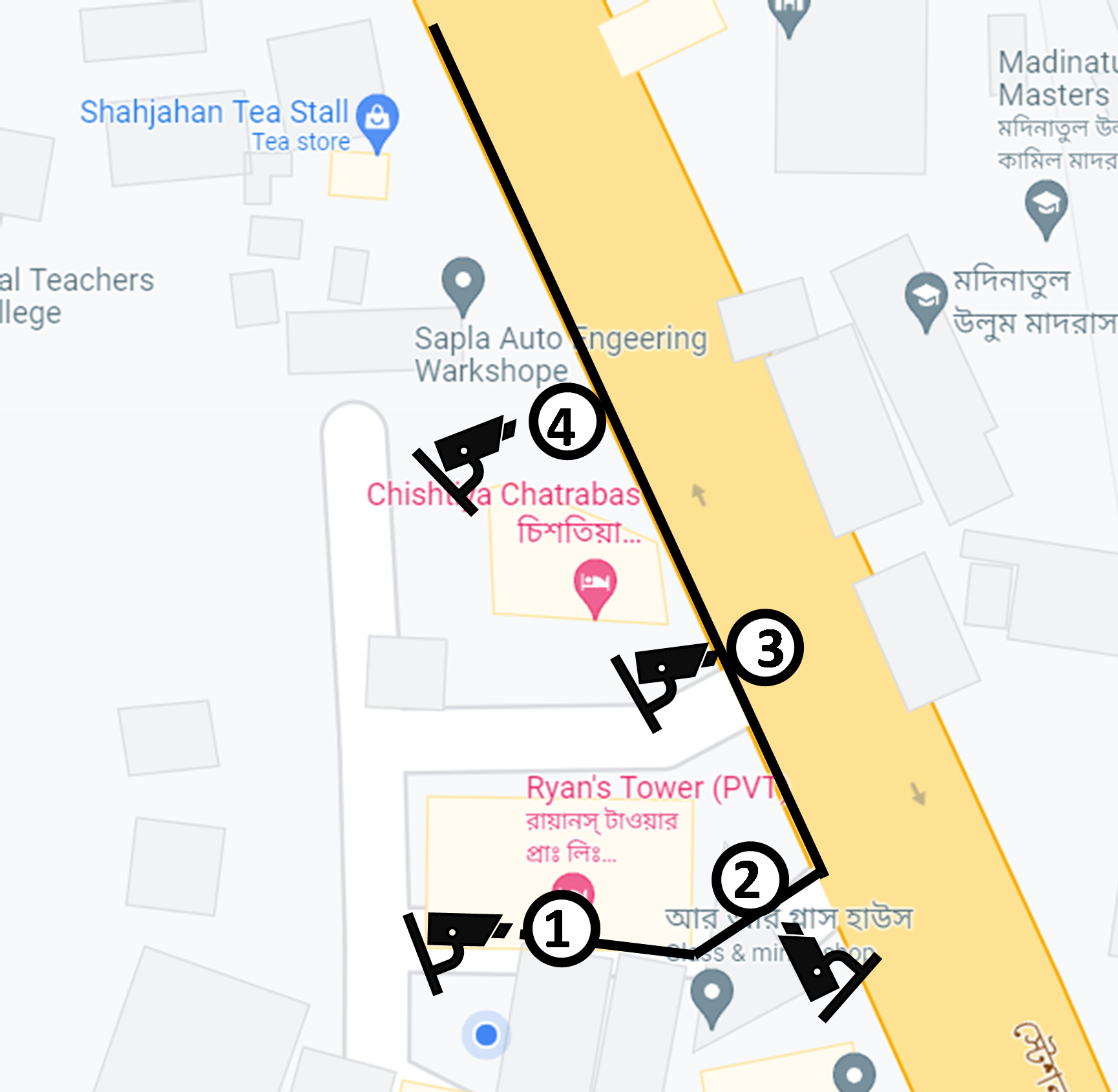}
\caption{Map of Surveillance Camera Setup}
\end{figure}

Figure 3 presents a map of the surveillance camera setup, illustrating the strategic placement of cameras for optimal coverage. We recorded video footage under natural conditions with participant consent, using high-quality cameras for clear side views. Each participant walked in both directions to ensure realism. The dataset includes recordings from four individuals to validate our methodology. 

\subsection{Dataset Description}
To obtain clear and accurate gait information, we strategically positioned cameras to ensure non-overlapping fields of view, capturing unique side-view perspectives of individuals. The footage was trimmed to include only segments where the full body was visible, maximizing data efficiency. Our dataset, created in an uncontrolled environment, provides a realistic representation of gait patterns, capturing data in various phases of the gait cycle for robust person identification and re-identification. We used a Samsung SM-S908E device with a bit rate of 20.0 Mb/s, a resolution of 1920 x 1080 pixels, a frame rate of 60 FPS, a bit depth of 8 bits, and a YUV color space. The dataset specifications are as follows: it includes 64 clips with a resolution of 1920 x 1080, a total size of 682 MB, and variations in angle, background, and view, utilizing four cameras with a frame rate of 60 FPS. This setup contrasts with the CASIA-B dataset \cite{iaCenterBiometrics}, which is collected in a controlled environment. The primary strengths of our dataset include its reflection of real-world scenarios, lighting variation, outdoor, complex environment and the potential for innovative applications. However, challenges include variability in walking patterns, potential biases, and varying data quality.

\section{Methodology}

The methodology consists of two main parts: Data Preprocessing and Classification Techniques. Data preprocessing is detailed first, outlining the sequence and logic of operations. This is followed by classification using various Machine Learning (ML) techniques, including ensemble learning, to process the created dataset. The aim is to extract key features from video data to train a classification model, ensuring robust and accurate results. Figure 4 depicts the flow diagram of the proposed methodology, outlining the key steps and processes involved in the approach.

\begin{figure}[h]
\centering
\includegraphics[width=0.4\textwidth]{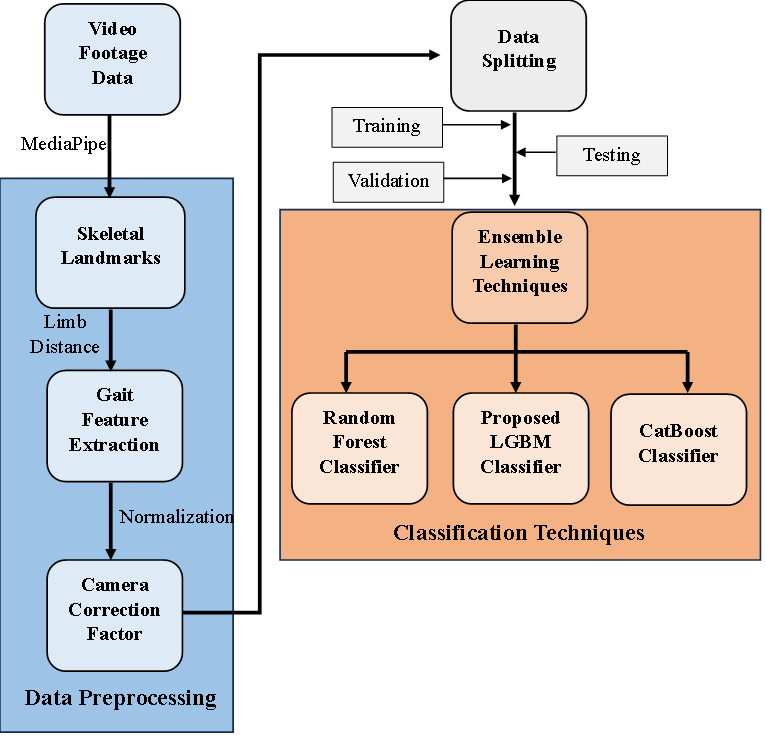}
\caption{The Flow Diagram of Proposed Methodology}
\end{figure}

\subsection{Data Preprocessing}
Data collected from uncontrolled environments requires processing before being used in our model. This involves identifying human body elements and motion from video footage. Processing starts when an individual enters the camera's field of view. Since the data comes from multiple non-overlapping cameras, correction factors for camera angles and views are necessary.

\subsubsection{Skeletal Landmarks}
Skeletal landmark extraction is crucial for posture estimation, with MediaPipe \cite{lugaresi2019mediapipe} and OpenPose \cite{cao2017realtime} being the two main methods. MediaPipe is favored due to its top-down approach, which first detects individuals and then identifies key points, leveraging TensorFlow for streamlined processing. The landmark detection is formulated as a regression problem, where the network 
predicts a set of keypoint coordinates \( \mathbf{p} = (x_i, y_i) \) for each landmark 
\( i \), minimizing the loss function:

\begin{equation}
L = \sum_{i} \| \hat{\mathbf{p}}_i - \mathbf{p}_i \|^2
\end{equation}

where \( \hat{\mathbf{p}}_i \) represents the predicted keypoint and \( \mathbf{p}_i \) 
is the ground truth. Additionally, MediaPipe refines landmarks using an affine transformation:

\begin{equation}
\mathbf{p'} = \mathbf{R} \mathbf{p} + \mathbf{t}
\end{equation}

where \( \mathbf{R} \) is the rotation matrix and \( \mathbf{t} \) is the translation vector. 
Also, some backend equations are central to MediaPipe’s process for detecting, refining, and tracking human poses in real-time. MediaPipe Pose detects 33 key points (landmarks) on the human body by estimating heatmaps \( H \) for each landmark. These heatmaps predict the probability that a given point \( i \) is located at coordinates \( (x, y) \):

\begin{equation}
H_i(x, y) = \text{probability that point } i \text{ is at location } (x, y)
\end{equation}

To estimate the depth of each landmark, MediaPipe uses regression-based models to calculate the \( z \)-coordinate for each keypoint. The final 3D coordinates are derived by combining the 2D heatmap predictions with the predicted depth \( Z_i \):

\begin{equation}
(x, y, z) = \text{argmax}_{(x, y)} H_i(x, y) + Z_i
\end{equation}

where \( Z_i \) is the predicted depth value for the \( i \)-th landmark.

Once the keypoints are detected and their 3D positions are estimated, MediaPipe constructs the human body's skeleton by connecting the detected landmarks. The connections between the landmarks are derived using predefined adjacency matrices, represented as:

\begin{equation}
S = \{ (l_i, l_j) \, | \, l_i, l_j \in \text{landmarks} \}
\end{equation}

For maintaining temporal consistency and ensuring smooth motion tracking across frames, MediaPipe refines the pose solution using a kinematic smoothing approach. This technique blends the current frame's landmarks \( L_t \) with the previous frame's landmarks \( L_{t-1} \), where \( \alpha \) is a smoothing factor that balances the two:

\begin{equation}
L_t = \alpha L_{t-1} + (1 - \alpha) L_t
\end{equation}

This process ensures that the detected pose remains consistent and stable over time, reducing jitter and improving the overall tracking quality.

Conversely, OpenPose uses a bottom-up method, generating keypoints with spatial convolutions and linking them with Part Affinity Fields (PAFs), requiring complex post-processing \cite{Team_2023}. Let \( H \) represent the heatmaps of the body parts, and \( P \) represent the Part Affinity Fields, which are used to model the association between parts:

\begin{equation}
H_i(x, y) = \text{probability of body part } i \text{ at location } (x, y)
\end{equation}

The Part Affinity Fields are used to link detected keypoints by providing associations between body part pairs \( (l_i, l_j) \):

\begin{equation}
P_{i,j}(x, y) = \text{affinity score between parts } l_i \text{ and } l_j \text{ at location } (x, y)
\end{equation}

This method requires additional post-processing to resolve ambiguities and link the detected keypoints. MediaPipe's hierarchical, semantic approach and efficient post-processing make it a superior choice for extracting skeletal landmarks compared to OpenPose \cite{Team_2023}.

\subsubsection{Landmarks Selection}
Skeletal landmarks, essential for posture estimation, are extracted using MediaPipe \cite{lugaresi2019mediapipe}, which provides 32 landmarks for various applications like hand gesture and facial expression recognition. The dataset's landmarks are detailed in Table 1, focusing on key points essential for gait analysis. Not all landmarks are necessary, and some are omitted to simplify the model and enhance identification accuracy. The data is recorded in CSV format with 64 videos for 4 individuals, resulting in 23 landmarks across 9,974 frames. MediaPipe processes each frame individually, discarding frames where landmarks are not detected, leading to a dataset of non-sequential frames without inherent temporal relationships.
\begin{table}[h]
\centering
\label{table}
\caption{Considered Landmarks}
\begin{tabular}{|c|c|}
\hline
\textbf{Upper Body } & \textbf{Lower   Body}  \\
\hline
LEFT\_EAR\_X       & LEFT\_HEEL\_X         \\
\hline
LEFT\_EAR\_Y       & LEFT\_HEEL\_Y         \\ 
\hline 
LEFT\_ELBOW\_X     & LEFT\_HIP\_X          \\
\hline
LEFT\_ELBOW\_Y     & LEFT\_HIP\_Y          \\ 
\hline 
LEFT\_WRIST\_X     & LEFT\_KNEE\_X         \\
\hline
LEFT\_WRIST\_Y     & LEFT\_KNEE\_Y         \\
\hline
RIGHT\_SHOULDER\_X & LEFT\_ANKLE\_X        \\
\hline
RIGHT\_SHOULDER\_Y & LEFT\_ANKLE\_Y        \\
\hline
                   & LEFT\_HEEL\_X         \\
\hline
                   & RIGHT\_HEEL\_X        \\
\hline
                   & RIGHT\_FOOT\_INDEX\_X \\
\hline
                   & RIGHT\_HIP\_X         \\
\hline 
                   & RIGHT\_HIP\_Y         \\
\hline
\end{tabular}                   
\end{table}

\subsubsection{Gait Feature Extraction}
To identify distinctive gait characteristics, we calculate several features from skeletal landmarks using Euclidean distances. These features include:

\begin{equation}
\text { distance }=\sqrt{(x 2-x 1)^2+(y 2-y 1)^2}
\end{equation}

\begin{enumerate}
    \item \textbf{Height}: Calculated as the distance between the ear and heel.
   \begin{equation}
\text{Height} = \sqrt{ 
\begin{aligned}
&\left( \text{LEFT}_{\text{EAR}_X} - \text{LEFT}_{\text{HEEL}_X} \right)^2 + \\
&\left( \text{LEFT}_{\text{EAR}_Y} - \text{LEFT}_{\text{HEEL}_Y} \right)^2 
\end{aligned}
}
\end{equation}

    \item \textbf{Hand Length}: The sum of the upper and lower hand lengths, calculated separately.
   
\begin{equation}
\text{Upper Hand} = \sqrt{ 
\begin{aligned}
&\left( \text{LEFT}_{\text{SHOULDER}_X} - \text{LEFT}_{\text{ELBOW}_X} \right)^2 + \\
&\left( \text{LEFT}_{\text{SHOULDER}_Y} - \text{LEFT}_{\text{ELBOW}_Y} \right)^2 
\end{aligned}
}
\end{equation}

\begin{equation}
\text{Lower Hand} = \sqrt{
\begin{aligned}
&\left( \text{LEFT}_{\text{ELBOW}_X} - \text{LEFT}_{\text{WRIST}_X} \right)^2 + \\
&\left( \text{LEFT}_{\text{ELBOW}_Y} - \text{LEFT}_{\text{WRIST}_Y} \right)^2
\end{aligned}
}
\end{equation}

    \begin{equation}
    \text{Hand} = \text{Upper Hand} + \text{Lower Hand}
    \end{equation}

    \item \textbf{Leg Length}: The sum of thigh and lower leg lengths.

\begin{equation}
\text{Thigh} = \sqrt{
\begin{aligned}
&\left( \text{LEFT}_{\text{HIP}_X} - \text{LEFT}_{\text{KNEE}_X} \right)^2 + \\
&\left( \text{LEFT}_{\text{HIP}_Y} - \text{LEFT}_{\text{KNEE}_Y} \right)^2
\end{aligned}
}
\end{equation}

\begin{equation}
\text{Lower Leg} = \sqrt{
\begin{aligned}
&\left( \text{LEFT}_{\text{KNEE}_X} - \text{LEFT}_{\text{ANKLE}_X} \right)^2 + \\
&\left( \text{LEFT}_{\text{KNEE}_Y} - \text{LEFT}_{\text{ANKLE}_Y} \right)^2
\end{aligned}
}
\end{equation}

\begin{equation}
\text{Leg} = \text{Thigh} + \text{Lower Leg}
\end{equation}
    \item \textbf{Step Length}: Measured between the heels of opposite feet.
   \begin{equation}
\text{Step Length} = \sqrt{
\begin{aligned}
&\left( \text{LEFT}_{\text{HEEL}_X} - \text{RIGHT}_{\text{HEEL}_X} \right)^2
\end{aligned}
}
\end{equation}

    \item \textbf{Foot Clearance}: Distance from one foot’s heel to the other foot's index.

\begin{equation}
\text{Foot Clearance} = \sqrt{
\begin{aligned}
&\left( \text{LEFT}_{\text{HEEL}_X}- \text{RIGHT}_{\text{FOOT}_{\text{INDEX}_X}} \right)^2
\end{aligned}
}
\end{equation}

    \item \textbf{Body Wideness}: Ratio of shoulder width to hip width.
\begin{equation}
\text{Hip Wideness} = \sqrt{
\begin{aligned}
&\left( \text{LEFT}_{\text{HIP}_X} - \text{RIGHT}_{\text{HIP}_X} \right)^2 + \\
&\left( \text{LEFT}_{\text{HIP}_Y} - \text{RIGHT}_{\text{HIP}_Y} \right)^2
\end{aligned}
}
\end{equation}

\begin{equation}
\text{Shoulder Wideness} = \sqrt{
\begin{aligned}
&\left( \text{LEFT}_{\text{S}_X} - \text{RIGHT}_{\text{S}_X} \right)^2 + \\
&\left( \text{LEFT}_{\text{S}_Y} - \text{RIGHT}_{\text{S}_Y} \right)^2
\end{aligned}
}
\end{equation}

\begin{equation}
\text{Body Wideness} = \frac{\text{Shoulder Wideness}}{\text{Hip Wideness}}
\end{equation}

    \item \textbf{Shoulder-Hip Ratio (SHR)}: Ratio of shoulder width to hip width.
    \begin{equation}
    \text{SHR} = \frac{\text{Shoulder Wideness}}{\text{Hip Wideness}}
    \end{equation}
\end{enumerate}

These features are normalized to a standard range for computational efficiency, facilitating the use of lightweight machine learning models. Note that the features derived are based solely on spatial data from individual frames, without considering temporal relationships, which are important for features like walking speed and cadence. This approach reduces storage needs and computational costs while enhancing model training and real-time identification capabilities.

\subsubsection{Camera Correction Factor}

\begin{figure}[h]
\centering
\includegraphics[width=0.5\textwidth]{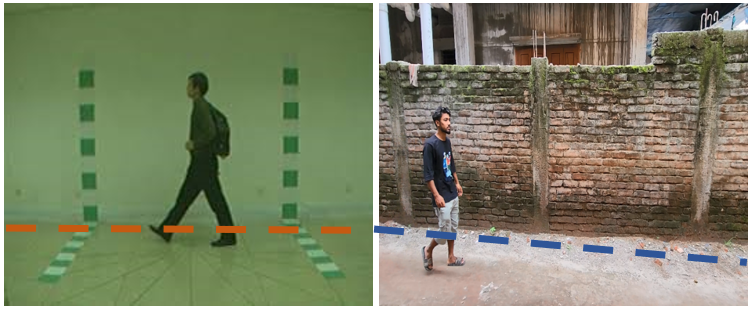}
\caption{Camera to Subject Distance Comparison with CASIA-B}
\label{fig:camera_comparison}
\end{figure}

Our dataset features various camera angles, heights, natural lighting conditions, and complex backgrounds. Unlike the CASIA-B dataset \cite{yaprak2024different}, where fixed distances and controlled backgrounds were used, our setup lacks such constraints. As shown in Figure \ref{fig:camera_comparison}, individuals in our dataset may walk in less controlled environments, leading to deviations in the camera-subject distance.

Figure \ref{fig:camera_comparison} illustrates that the person on the right does not follow a linear path compared to the person on the left. This deviation arises from varying camera distances. To address this, we apply a correction factor to standardize frame-based features across different cameras.

To ensure consistency in height measurements across multiple cameras, we establish the height from Camera 1 as a reference point. Deviations observed in the height measurements from other cameras are then quantified using a correction factor, which is defined as follows:
\begin{equation}
\begin{aligned}
    H_{PC} &= \frac{\sum_{i=0}^{N_C}h_{i_C}}{N_C} \\
    \text{Camera Correction Factor} &= \frac{H_{PC}}{H_{P1}}
\end{aligned}
\end{equation}

Here, \(P\) and \(C\) denote the person and camera numbers, respectively. \(h_{iC}\) is the height of person \(P\) in frame \(i\) for camera \(C\). \(H_{PC}\) is the average height of person \(P\) in camera \(C\), and \(H_{P1}\) is the average height of person \(P\) in camera 1. The calculated correction factors are shown in Table \ref{table:correction_factors}.

\begin{table}[h]
\centering
\caption{Correction Factor of Cameras}
\label{table:correction_factors}
\begin{tabular}{|c|c|}
\hline
\textbf{Camera No.} & \textbf{Correction Factors} \\
\hline
1 & 1.000000 \\
\hline
2 & 1.023697 \\
\hline
3 & 1.779473 \\
\hline
4 & 1.562166 \\
\hline
\end{tabular}
\end{table}

The correction factors are applied to all features, excluding Person ID and Frame number, to minimize errors effectively.

The summary of the data of limited features are shown in Figure 4.5 for ease in representation.

The processed data are visualized to ensure quality. Figure 6 shows the distribution of data per person. It is evident that the data availability varies slightly across classes, which is justifiable. For example, Person 3, who walks faster, has a shorter video duration.
\begin{figure}[h]
\centering
\includegraphics[clip, trim=0cm 16cm 0cm 0cm, width=0.5\textwidth]{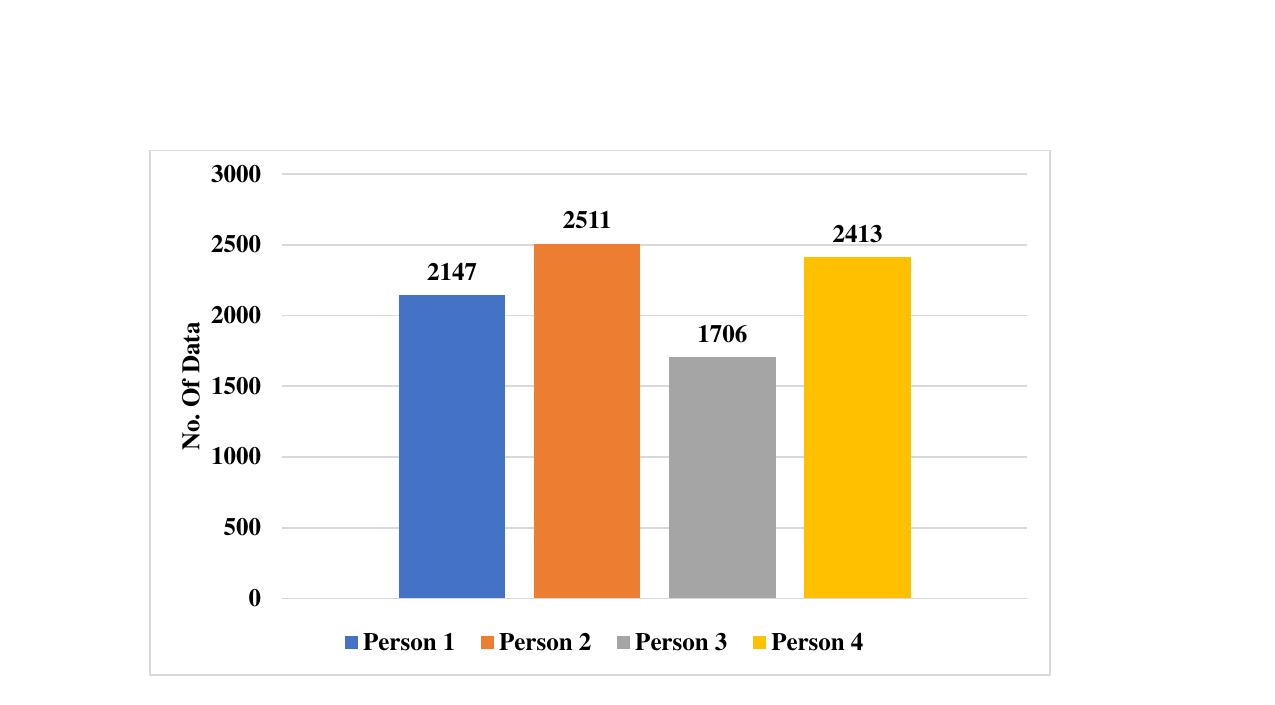}
\caption{Data Per Class}
\end{figure}

To evaluate data quality, gait features are compared across persons. For instance, height correlates at 0.69 with upper hand length and 0.94 with leg length. A positive correlation of 0.72 is observed between step length and foot clearance. Understanding these relationships aids in improving model performance and classification.

\subsection{Data Splitting}
After extracting gait features, the data is split into training, validation, and testing sets to ensure reliable model evaluation and prevent overfitting. The dataset comprises 54 films with 8773 rows for training. Validation uses 25\% of the training data, amounting to 2194 records. Testing involves 8 films with 1201 rows. In total, there are 9974 rows.

\subsection{Classification Techniques}
In classification, both Machine Learning (ML) and Deep Learning (DL) methodologies are used to categorize data. ML techniques, including Ensemble Learning, offer lower computational complexity and faster training compared to DL methods like Long Short-Term Memory (LSTM). For our non-sequential data, Ensemble Learning provides satisfactory performance and efficiency, aligning with recent studies \cite{saha2022demand} which suggest it outperforms LSTM in numerical data processing. Our goal is to enhance training efficiency by reducing computational complexity and data storage needs. 

The choice of the LGBM (LightGBM) model for gait classification in this research is driven by its significant advantages over not only ensemble methods such as Random Forest and CatBoost \cite{zhou2020ensemble}, but also traditional machine learning and deep learning models \cite{ke2017lightgbm}. LGBM’s gradient boosting framework efficiently aggregates weak learners to form a highly accurate model, making it an ideal solution for complex tasks like person identification and re-identification in resource-constrained settings \cite{chen2016xgboost}.

In comparison to deep learning approaches such as convolutional neural networks (CNNs) and recurrent neural networks (RNNs), which demand substantial computational resources and memory \cite{khan2021deep}, LGBM provides a faster and more efficient alternative. While deep learning excels in feature extraction, its high computational cost makes it unsuitable for environments where minimal memory usage is critical \cite{zhou2019gait}. Additionally, traditional models like support vector machines (SVMs) and k-nearest neighbors (k-NN) struggle with scalability and performance when faced with large, high-dimensional datasets like those in gait recognition \cite{zhang2018gait}. LGBM, by contrast, processes such data efficiently without sacrificing accuracy \cite{nishikawa2020gait}.

Moreover, simpler algorithms such as logistic regression and decision trees lack the complexity required to capture the subtle variations in gait patterns, particularly under challenging conditions like non-overlapping cameras and diverse walking styles \cite{wang2021gait}. LGBM’s leaf-wise growth strategy, along with its ability to effectively manage both continuous and categorical data, allows it to outperform in terms of accuracy and adaptability \cite{liu2019efficient}.

Compared to other ensemble methods like AdaBoost and XGBoost, LGBM is notably more efficient in memory usage and training speed \cite{li2019lightgbm}. Its advanced regularization techniques prevent overfitting \cite{shen2020regularization}, while its support for parallelism and optimized tree-learning algorithms result in faster convergence. Unlike XGBoost’s level-wise tree growth, LGBM’s leaf-wise approach reduces computation time, further enhancing performance \cite{ke2017lightgbm}.

In summary, LGBM surpasses a wide range of models by offering an unparalleled combination of speed, accuracy, and efficiency, making it the most suitable choice for developing a real-world, low-cost, memory-efficient gait-based re-identification system.

Ensemble Learning combines multiple models to enhance predictive performance, robustness, and generalization. By leveraging model diversity, ensemble methods often achieve better accuracy than individual models. Key ensemble algorithms include Random Forest (RF), Light Gradient Boosting Machine (LGBM), and CatBoost. We propose using the LGBM classifier for our data, and will demonstrate its superior performance compared to other ensemble techniques

\subsubsection{Proposed LGBM Classifier}
The Light Gradient Boosting Machine (LightGBM) is a powerful machine learning system for classification and regression, especially effective with large datasets and high-dimensional features. Its main strengths are speed and memory efficiency. LightGBM uses gradient boosting, which builds models iteratively to correct errors from previous rounds, leading to reliable predictions.

LightGBM employs histogram-based data splitting to construct decision trees leafwise, optimizing the classification process. Its lightweight architecture supports real-time predictions and handles categorical features without one-hot encoding, saving memory and simplifying preprocessing. It also includes regularization to prevent overfitting and uses advanced techniques like Gradient-based One Side Sampling (GOSS) and Exclusive Feature Bundling (EFB).

\begin{figure}[h]
\centering
\includegraphics[width=0.5\textwidth]{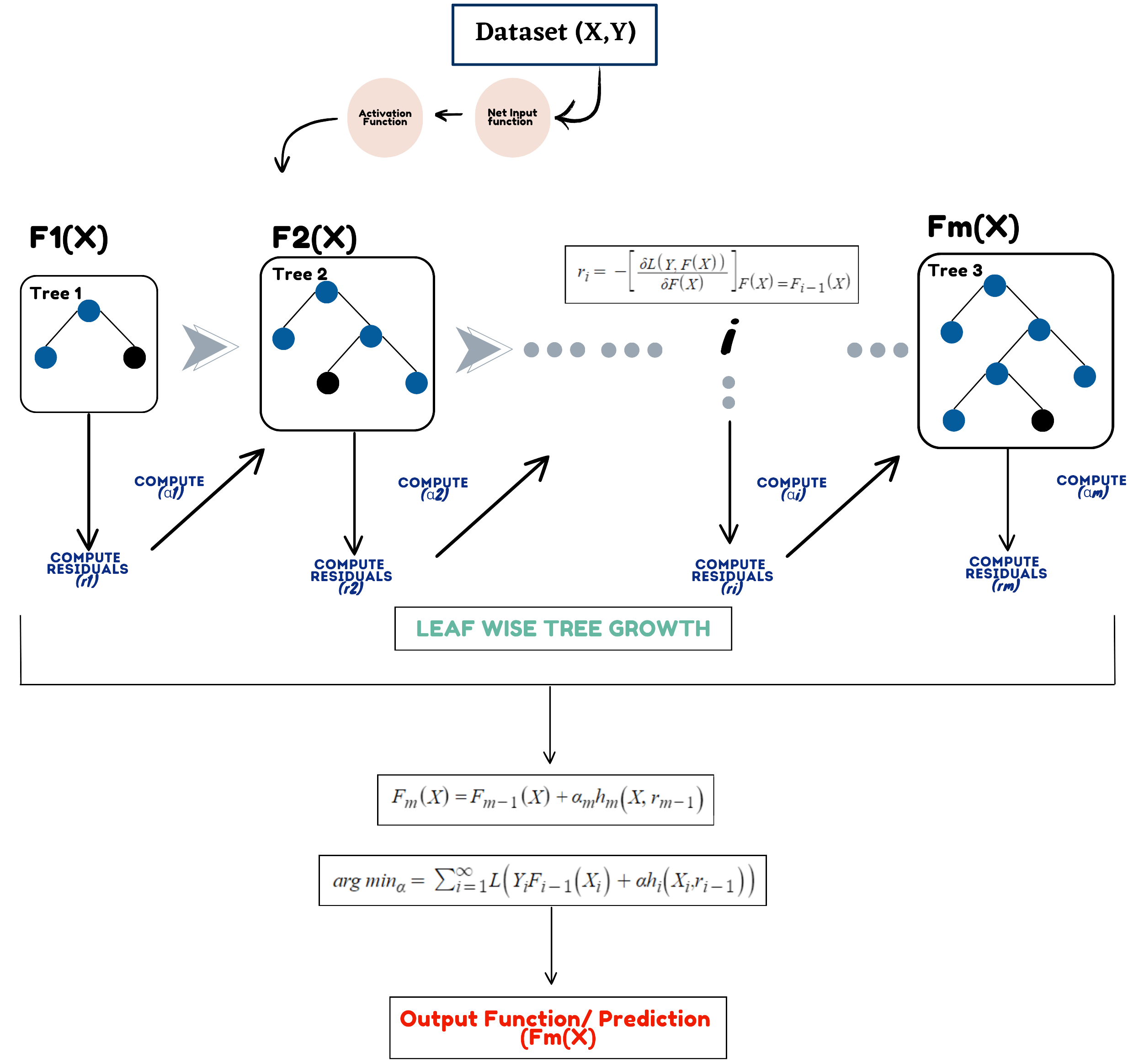}
\caption{LGBM Classifier}
\end{figure}
Figure 7 illustrates the LGBM Classifier, showcasing its structure and functionality within the context of the proposed model.
The GOSS technique enhances variance gain calculations by prioritizing instances with large gradients and normalizing gradient sums. The formula for variance gain is:

\begin{equation}
    Y^{\prime}(e) = \frac{1}{n} \sqrt{ 
    \begin{aligned}
    &\frac{\left(\sum_{x_{i}\in A_{l}} h_{i} + \frac{1-p}{q} \sum_{x_{i}\in B_{l}} h_{i}\right)^{2}}{n_{i}^{j}(e)} + \\
    &\frac{\left(\sum_{x_{i}\in A_{r}} h_{i} + \frac{1-p}{q} \sum_{x_{i}\in B_{r}} h_{i}\right)^{2}}{n_{i}^{j}(e)}
    \end{aligned}
    }
\end{equation}

\begin{equation}
    \begin{aligned}
    A_{l} &= \{x_{i} \in A \mid x_{i j} \leq e\}, \\
    A_{r} &= \{x_{i} \in A \mid x_{i j} > e\}, \\
    B_{l} &= \{x_{i} \in B \mid x_{i j} \leq e\}, \\
    B_{r} &= \{x_{i} \in B \mid x_{i j} > e\}
    \end{aligned}
\end{equation}

The GOSS (Gradient-based One-Side Sampling) technique starts with gradient boosting on a training set of $n$ instances $(x_1, \dots, x_n)$ in space $X_s$. In each boosting iteration, the negative gradients of the loss function are denoted as $(h_1, \dots, h_n)$. The top-a 100\% instances with the largest gradients form subset $A$, after sorting instances by their absolute gradient values \cite{sheridan2021light}. Instances are then divided according to predicted variance gain at vector $Y_{j}^{\prime}(d)$ over subset $B$, while the remaining $(1-p)$ 100\% of instances form subset $A$. The sum of gradients across $B$ is normalized to $A$ using the coefficient $(1-p)/q$.

In terms of Exclusive Feature Bundling (EFB), high-dimensional, sparse data can benefit from combining mutually incompatible features into exclusive feature bundles. This reduces the number of features without losing information, shifting histogram complexity from $O(\text{data} \times \text{feature})$ to $O(\text{data} \times \text{bundle})$, where $\text{bundle} << \text{feature}$, increasing training speed without sacrificing accuracy. The EFB technique reduces feature space by bundling mutually exclusive features, which speeds up histogram creation and training without sacrificing accuracy.

\begin{figure}[h]
\centering
\includegraphics[width=0.5\textwidth]{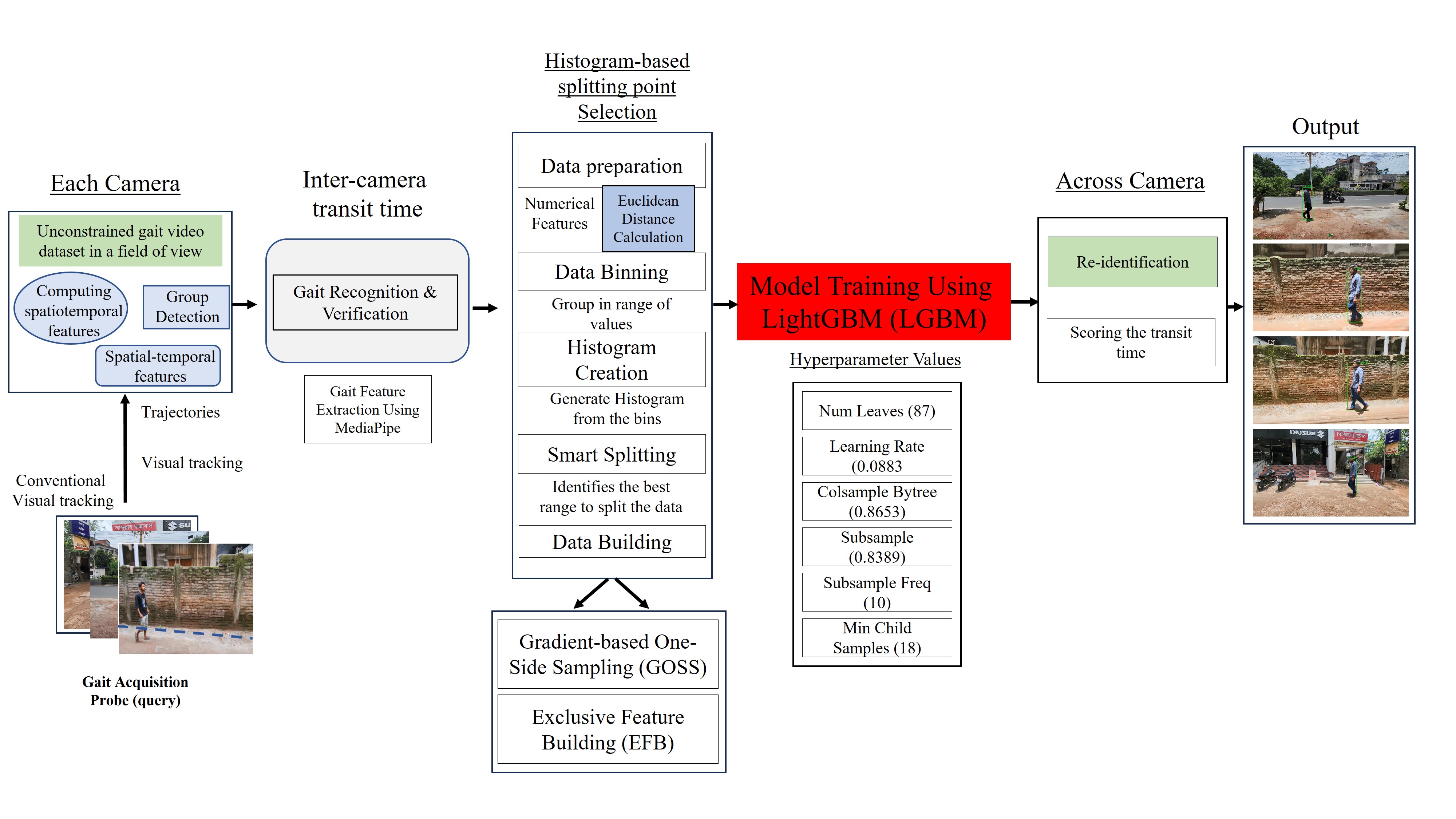}
\caption{Methodology}
\end{figure}
Figure 8  presents the overall framework utilized in this research, outlining the key processes and stages involved in developing the proposed model.
Hyperparameter tuning was performed using Optuna \cite{OptunaContributors2018}, a Python library for optimization. It utilizes algorithms like Bayesian optimization and Tree-structured Parzen Estimators (TPE) to find the best hyperparameters. For our case, TPE was used.

\begin{table}[h]
\centering
\caption{Hyperparameter Values}
\begin{tabular}{|c|c|}
\hline
\textbf{Hyperparameters} & \textbf{Value} \\
\hline
Num Leaves & 87 \\
\hline
Learning Rate & 0.0883 \\
\hline
Colsample Bytree & 0.8652 \\
\hline
Subsample & 0.8389 \\
\hline
Subsample Freq & 10 \\
\hline
Min Child Samples & 18 \\
\hline
\end{tabular}
\end{table}

Table 3 lists hyperparameters used for the LGBM model: `num leaves` (87) for tree construction, `learning rate` (0.0883) for model optimization, `colsample bytree` (0.8652) for feature diversity, `subsample` (0.8389) for training data fraction, `subsample freq` (10) for stability, and `min child samples` (18) to control leaf size. These settings ensure a well-balanced model that generalizes effectively from the data.

\section{Result Analysis and Discussion}
This section analyzes the performance our proposed LGBM model. It is divided into two main sections: performance matrix evaluation and visual classification. The performance matrix section compares model classifications based on gait feature data, while the visual classification section provides visual outputs for identification and reclassification.

\subsection{Performance Metrics}

\begin{enumerate}
    \item \textbf{Accuracy:} \\
    Defined as the ratio of correctly predicted instances to the total instances, accuracy is expressed mathematically as:
    \[
    \text{Accuracy} = \frac{\text{TP} + \text{TN}}{\text{TP} + \text{FP} + \text{TN} + \text{FN}}
    \]
    where TP, TN, FP, and FN represent true positives, true negatives, false positives, and false negatives, respectively. While a straightforward metric, accuracy may not effectively evaluate models on imbalanced datasets.

    \begin{figure}[h]
        \centering
        \includegraphics[width=0.5\textwidth]{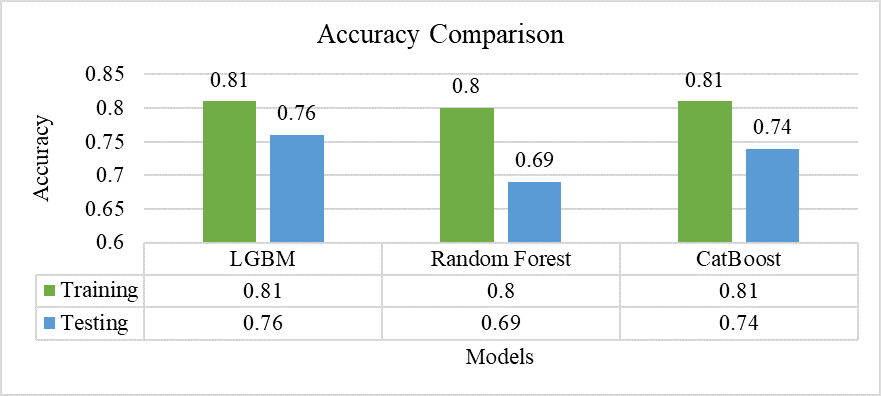}
        \caption{Accuracy Comparison: LGBM, Random Forest, CatBoost}
    \end{figure}

    \item \textbf{Precision:} \\
    Precision quantifies the accuracy of positive predictions and is calculated as:
    \[
    \text{Precision} = \frac{\text{TP}}{\text{TP} + \text{FP}}
    \]
    This metric is critical in contexts where false positives are particularly detrimental.

    \begin{figure}[h]
        \centering
        \includegraphics[width=0.5\textwidth]{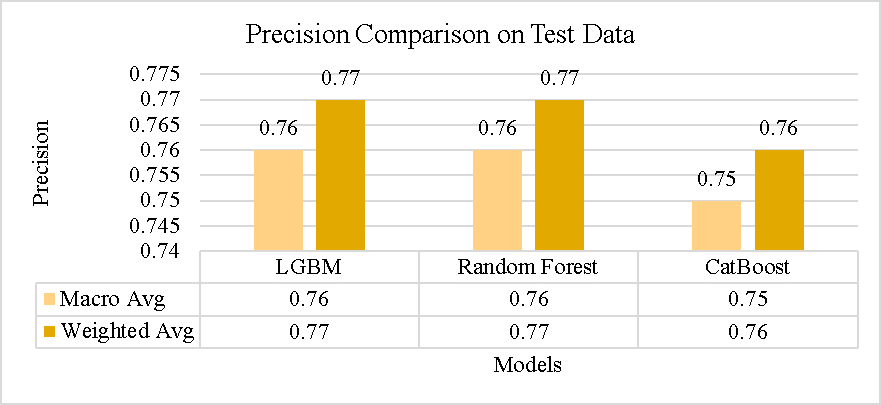}
        \caption{Precision Comparison: LGBM, Random Forest, CatBoost}
    \end{figure}

    \item \textbf{Recall:} \\
    Also known as sensitivity, recall measures the fraction of actual positives that are correctly identified:
    \[
    \text{Recall} = \frac{\text{TP}}{\text{TP} + \text{FN}}
    \]
    It is essential in applications where capturing all relevant instances is crucial.

    \begin{figure}[h]
        \centering
        \includegraphics[width=0.5\textwidth]{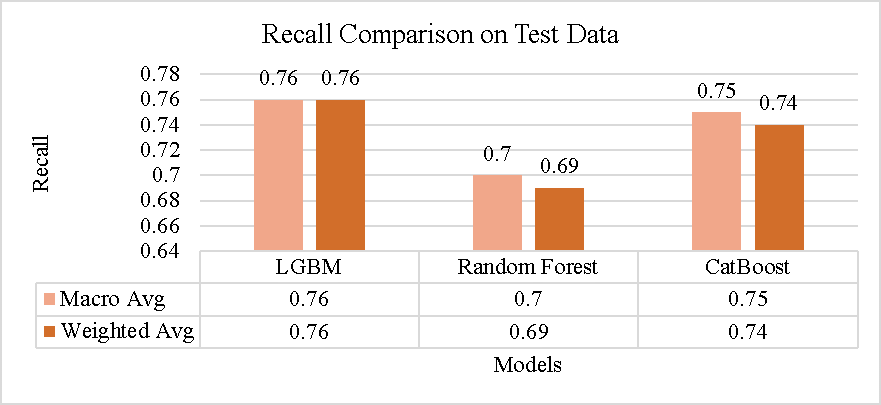}
        \caption{Recall Comparison: LGBM, Random Forest, CatBoost}
    \end{figure}

    \item \textbf{F1 Score:} \\
    The F1 score is the harmonic mean of precision and recall, given by:
    \[
    \text{F1 Score} = 2 \cdot \frac{\text{Precision} \times \text{Recall}}{\text{Precision} + \text{Recall}}
    \]
    This metric is particularly useful for evaluating models on imbalanced datasets.

    \begin{figure}[h]
        \centering
        \includegraphics[width=0.5\textwidth]{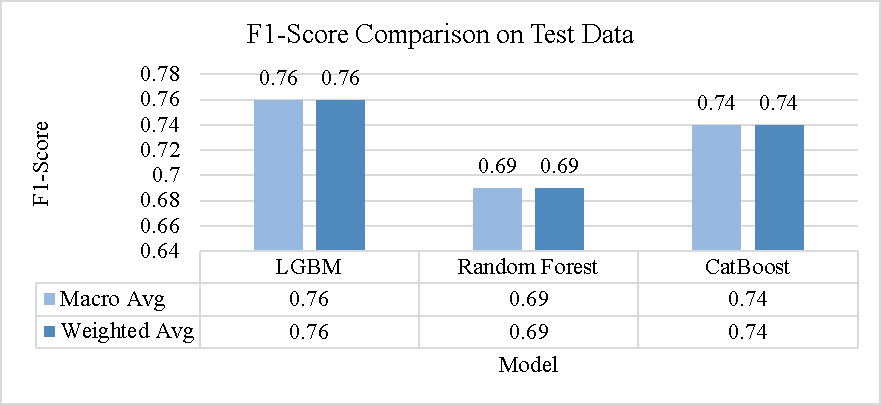}
        \caption{F1-Score Comparison: LGBM, Random Forest, CatBoost}
    \end{figure}

    \item \textbf{Confusion Matrix:} \\
    The confusion matrix in Figure 13 reveals that the custom LGBM model produced only one false positive in the fire class, indicating a strong classification performance reflected in high accuracy.

    \begin{figure}[h]
    \centering

    \begin{subfigure}{0.45\textwidth}
        \centering
        \includegraphics[width=\linewidth]{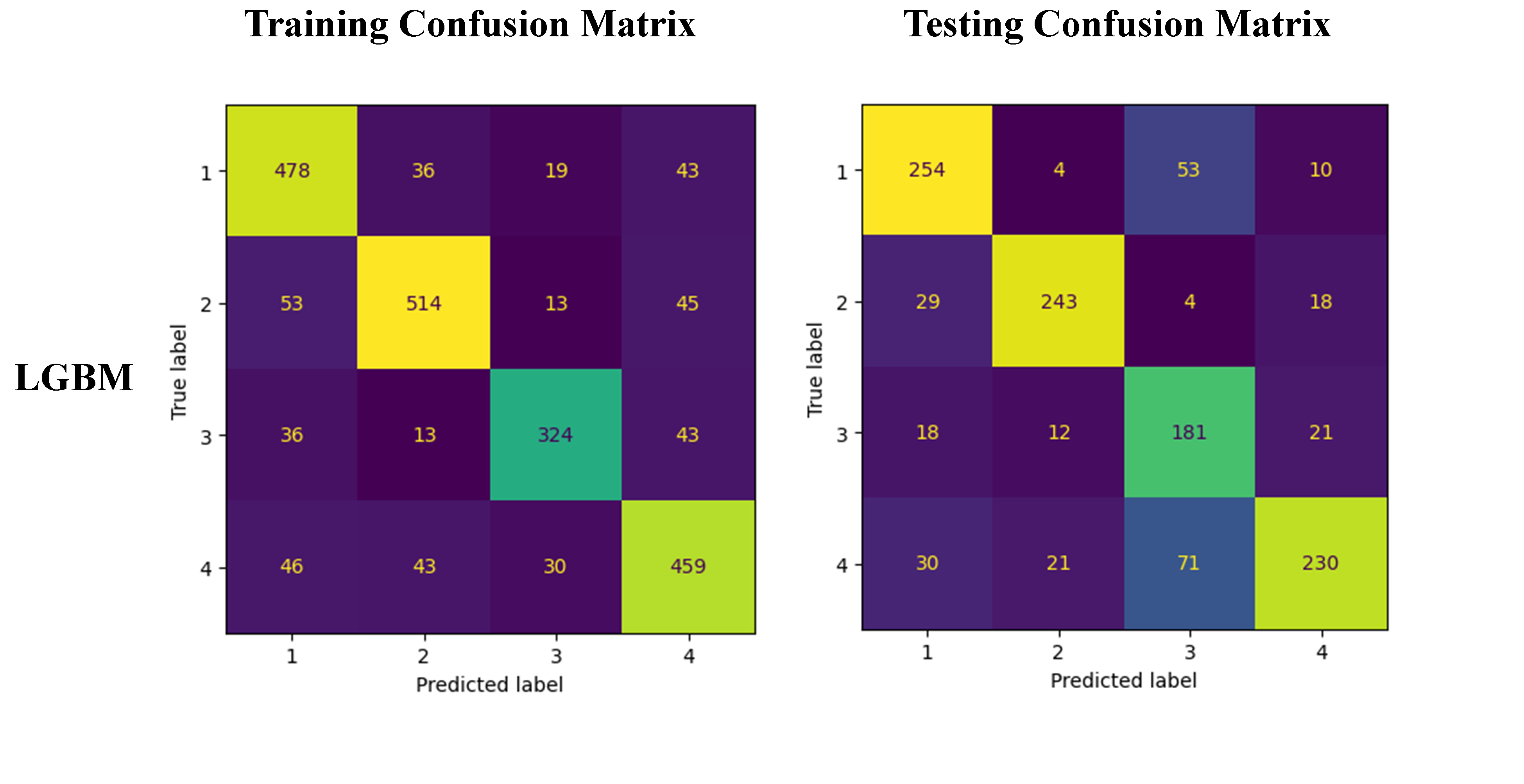}
        
    \end{subfigure}
    \hfill
    \begin{subfigure}{0.45\textwidth}
        \centering
        \includegraphics[width=\linewidth]{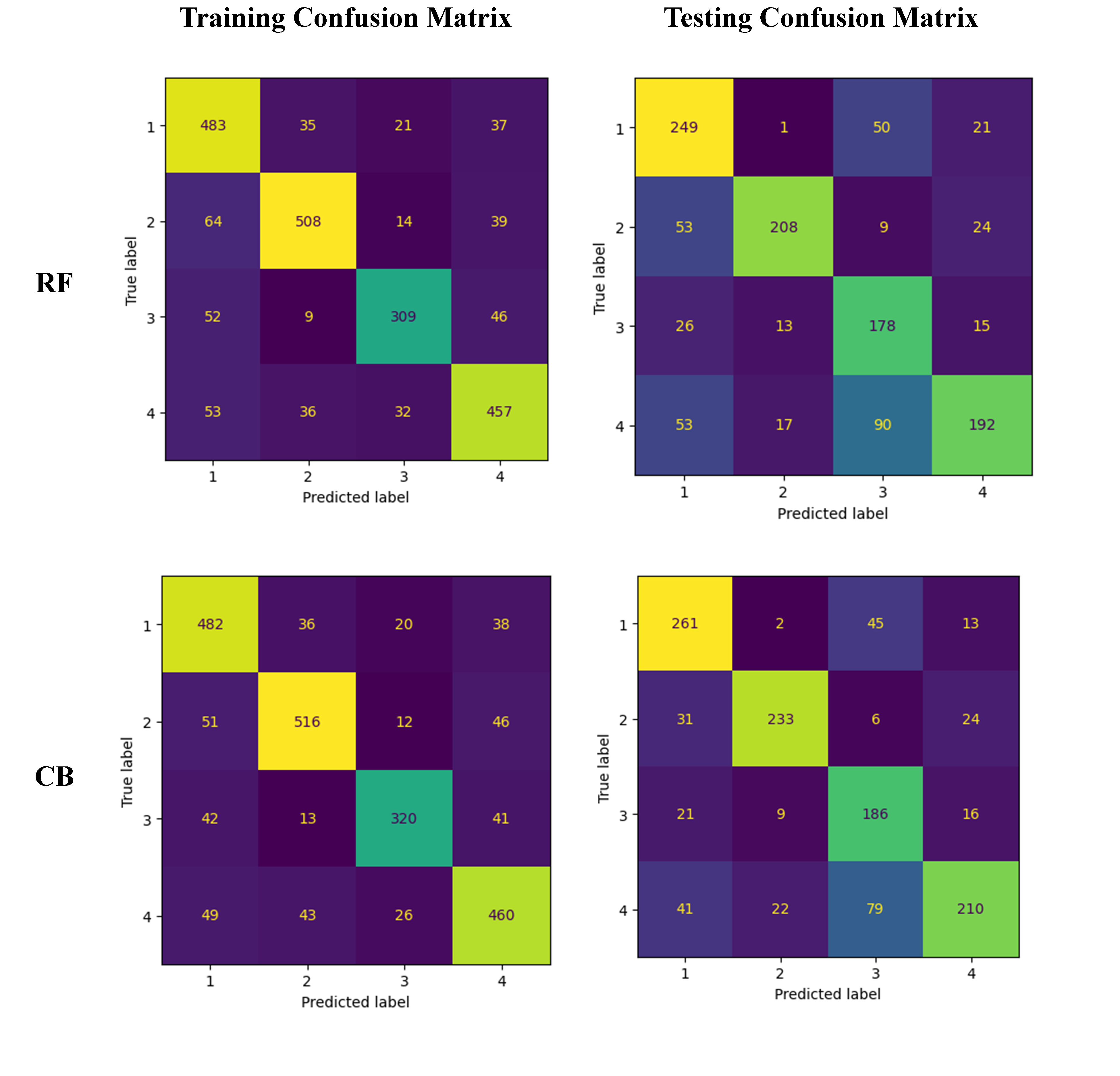}
       
    \end{subfigure}
    \caption{Confusion Matrix of LGBM, Random Forest, and CatBoost}
\end{figure}
\end{enumerate}
\subsection{Visual Classification}
Our system efficiently processes video data to identify and track individuals across multiple cameras using bounding boxes, ensuring accurate identification and re-identification. Figures 15(a) and 15(b) demonstrate the system’s ability to identify and track a person as they move from Camera 1 to Camera 2. Figure 15(c) shows the system's effectiveness in challenging conditions, such as poor lighting and complex backgrounds, highlighting its reliance on skeleton-based features rather than appearance. Figure 15(d) illustrates performance in varying lighting and distances, showing the model's robustness. Although some misclassifications occur, the method achieves around 76\% accuracy on unseen data, with maximum voting techniques addressing these issues.
\begin{figure}[h]
    \centering
    \begin{subfigure}{0.3\textwidth}
        \centering
        \includegraphics[width=\linewidth]{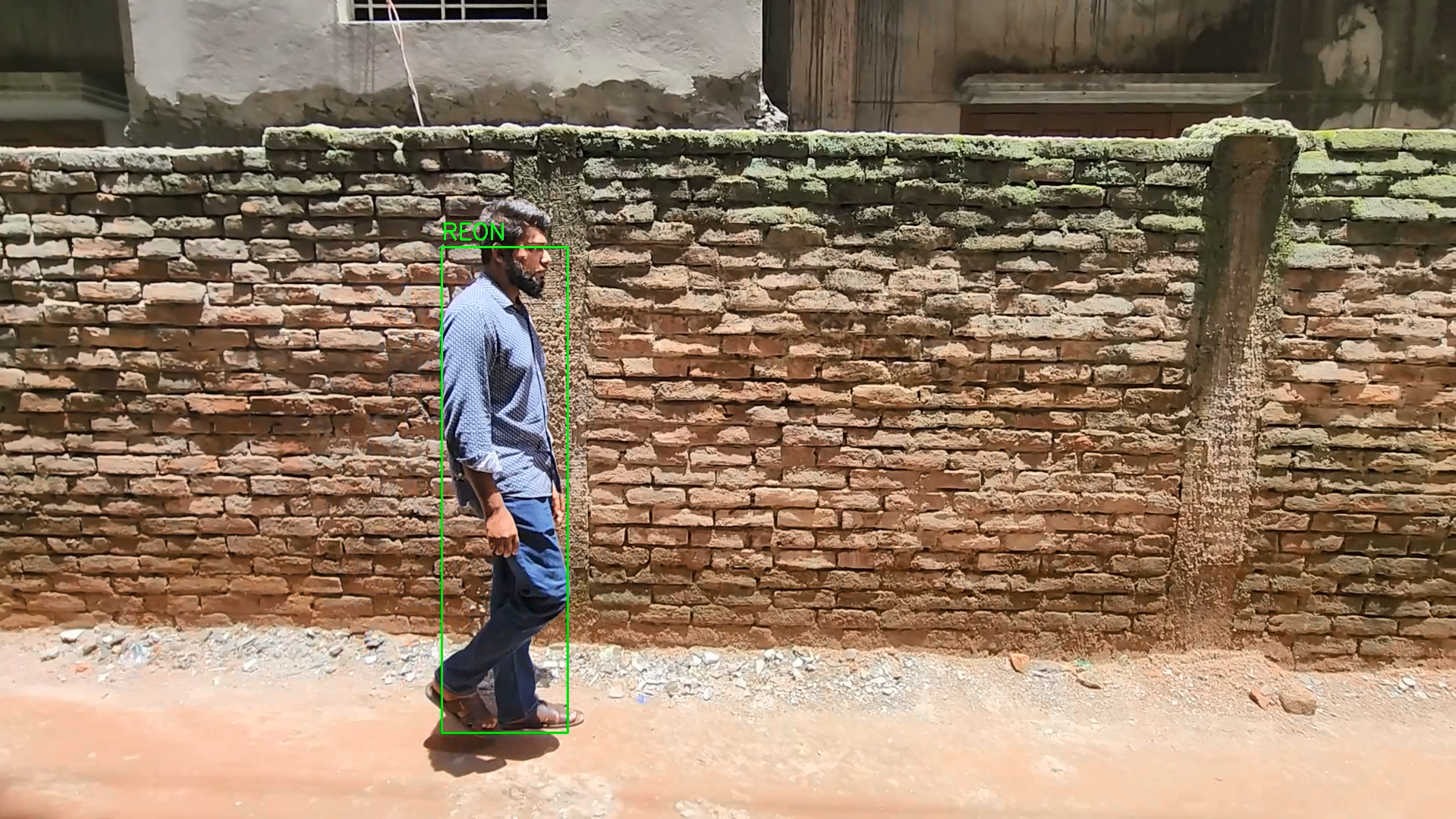}
        \caption{Person Identified in Camera 1}
    \end{subfigure}
    \hfill
    \begin{subfigure}{0.3\textwidth}
        \centering
        \includegraphics[width=\linewidth]{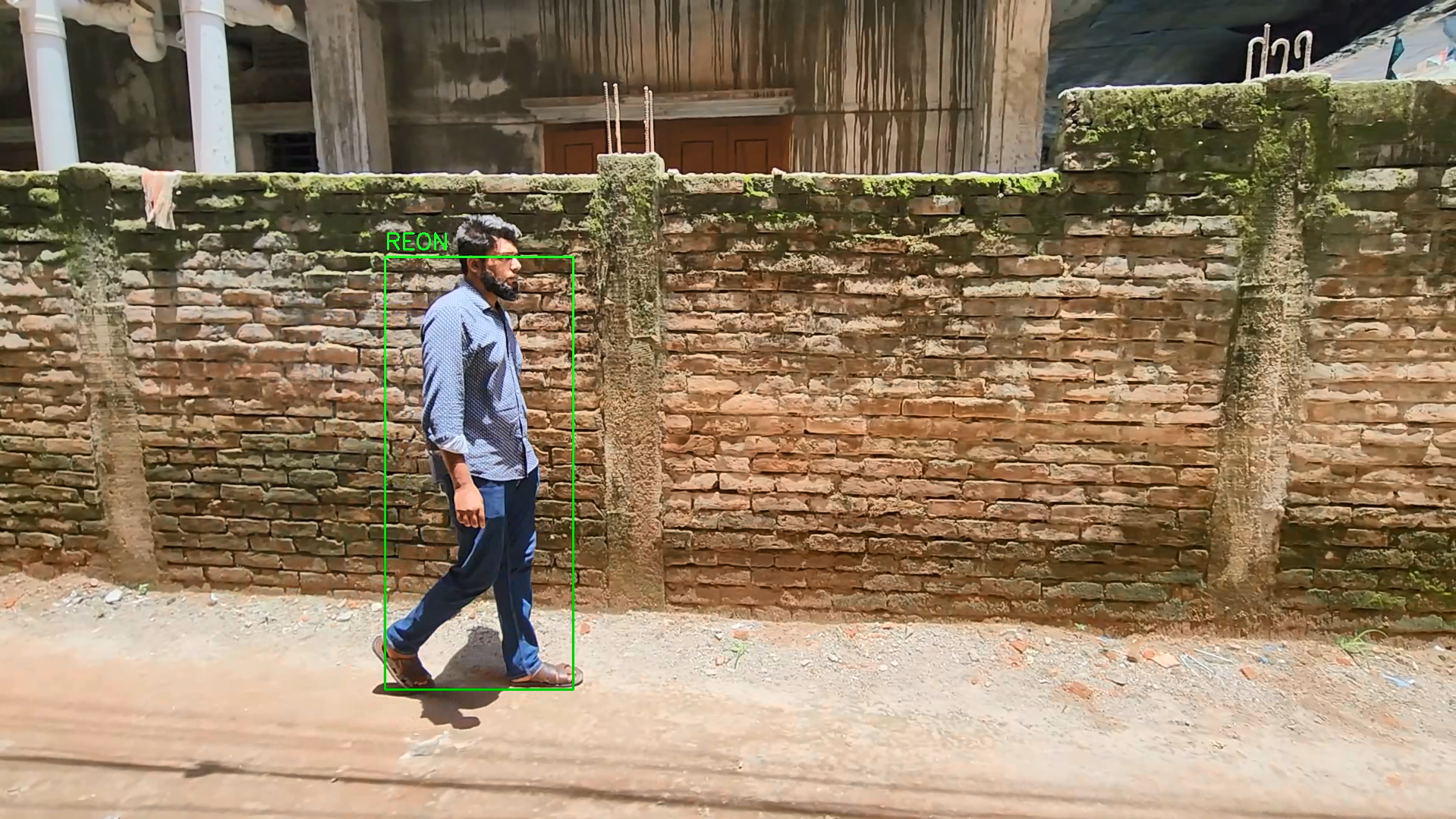}
        \caption{Person Re-identified in Camera 2}
    \end{subfigure}
\end{figure}

\begin{figure}[h]
    \centering
    \begin{subfigure}{0.3\textwidth}
        \centering
        \includegraphics[width=\linewidth]{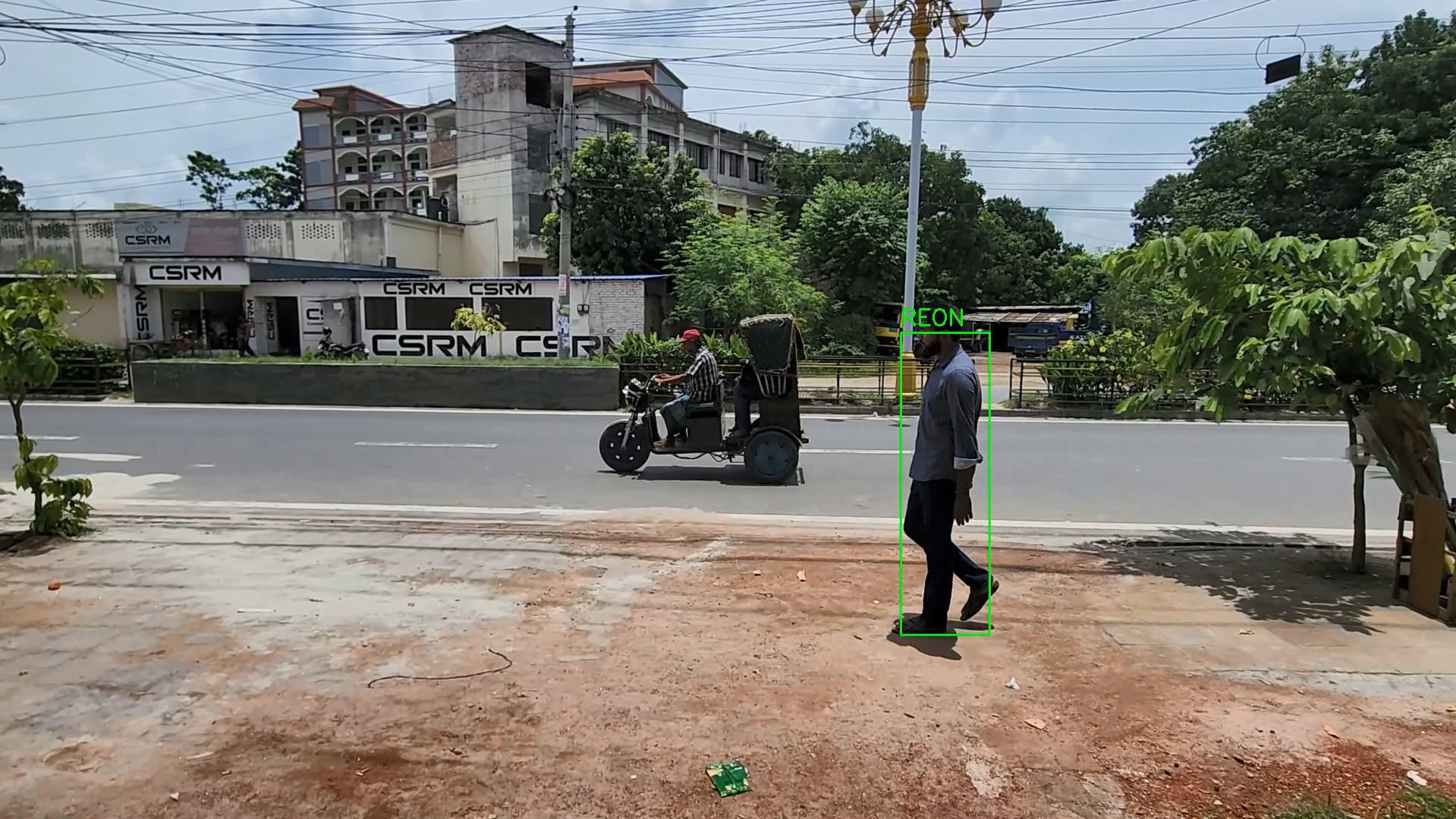}
        \caption{Person Re-identified in Camera 3}
    \end{subfigure}
    \hfill
    \begin{subfigure}{0.3\textwidth}
        \centering
        \includegraphics[width=\linewidth]{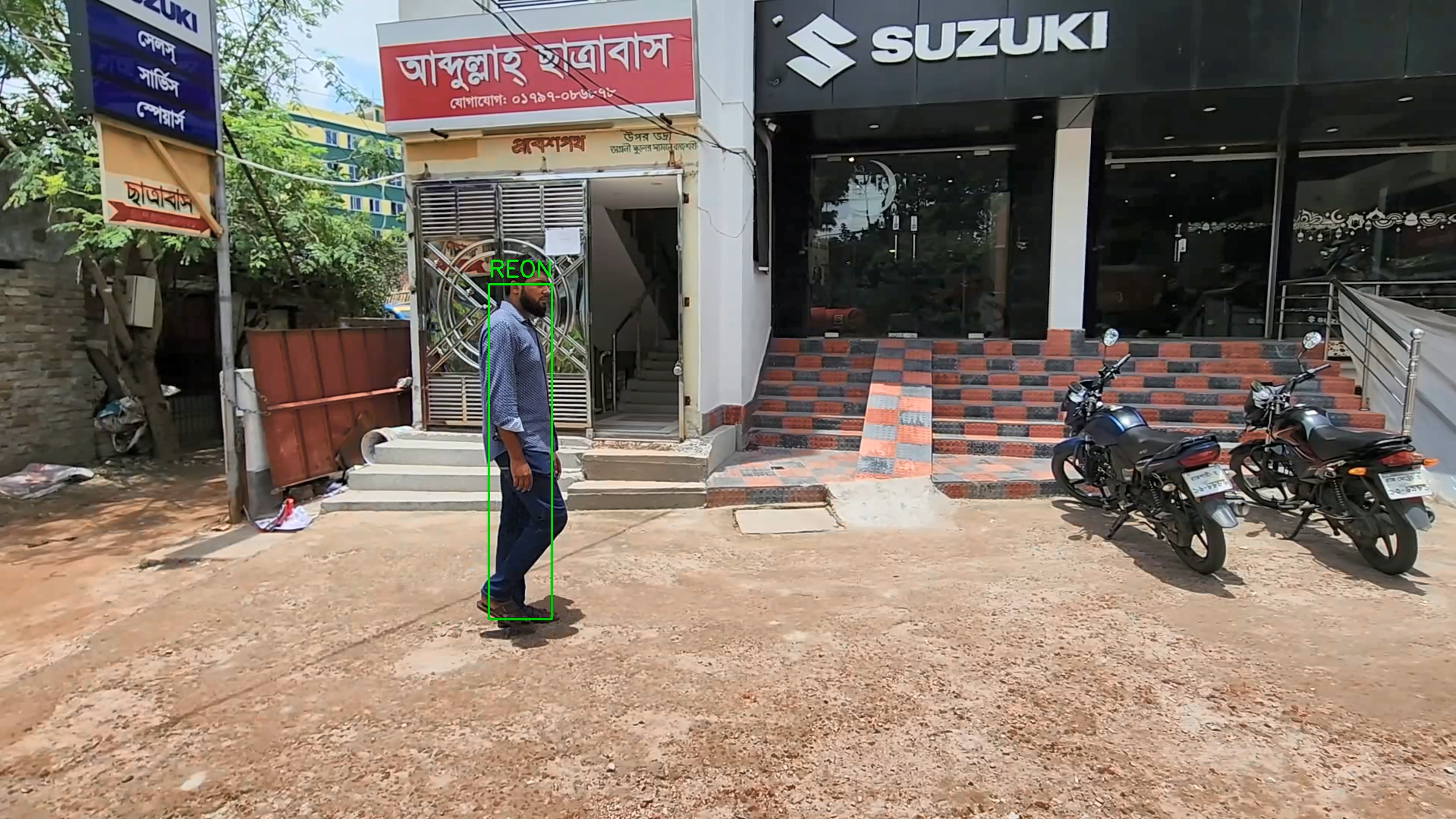}
        \caption{Person Re-identified in Camera 4}
    \end{subfigure}
    \caption{Person Identified and Re-identified in Camera 1,2,3,4}
\end{figure}

\subsection{Comparison of Computational Cost}

This section compares the computational cost, including training time and memory usage, of the LGBM, Random Forest, and CatBoost models. 

Figure 16 shows the training times for each model. The LGBM model is the fastest, requiring approximately 3.05 seconds, followed by Random Forest at 3.32 seconds. CatBoost has the longest training time at around 15.83 seconds, making LGBM preferable for tasks needing rapid training. Figure 17 illustrates the memory usage during training. The LGBM model used 554 MB, less than Random Forest (560 MB) and CatBoost (556 MB). This lower memory consumption highlights LGBM's efficiency, making it ideal for resource-constrained environments.

\begin{figure}[h]
\centering
\includegraphics[width=0.5\textwidth]{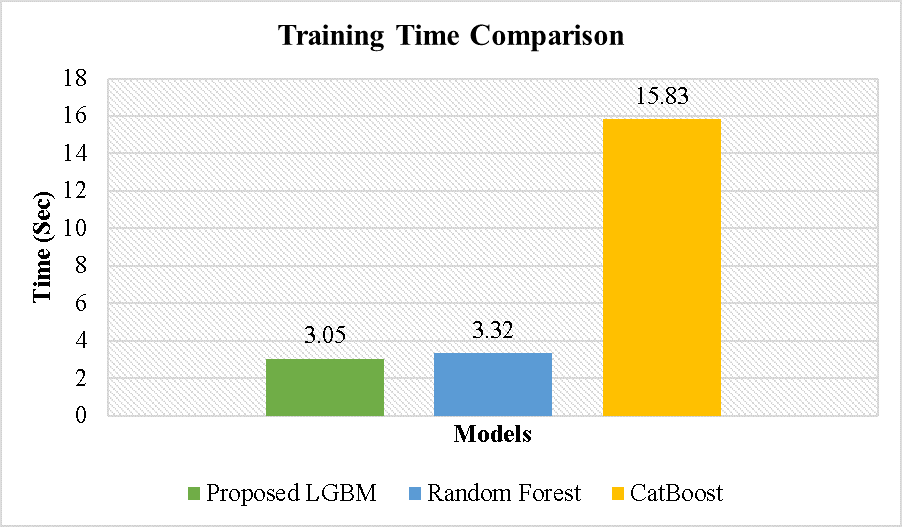}
\caption{Training Time Comparison Chart}
\end{figure}

\begin{figure}[h]
\centering
\includegraphics[width=0.5\textwidth]{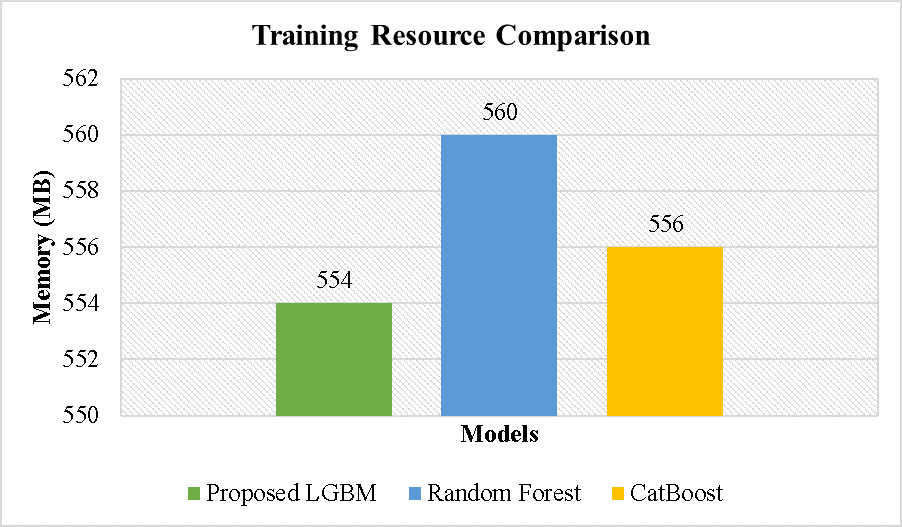}
\caption{Training Resources Comparison Chart}
\end{figure}

\begin{table}[h]
    \centering
    \caption{Comparison of Training and Testing Scores for Different Models}
    \begin{tabular}{|c|c|c|c|c|c|} 
        \hline
        \textbf{Model Name} & \textbf{Class} & \textbf{Precision} & \textbf{Recall} & \textbf{F1 Score} & \textbf{Support} \\ 
        \hline
        \multicolumn{6}{|c|}{\textbf{Training Scores}} \\ 
        \hline
        \multirow{4}{*}{\textbf{Proposed Model}} & 1 & 0.78 & 0.83 & 0.80 & 576 \\ \cline{2-6}
                                                & 2 & 0.85 & 0.82 & 0.84 & 625 \\ \cline{2-6}
                                                & 3 & 0.84 & 0.78 & 0.81 & 416 \\ \cline{2-6}
                                                & 4 & 0.78 & 0.79 & 0.79 & 578 \\ 
        \hline
        \multirow{4}{*}{\textbf{Random Forest}} & 1 & 0.74 & 0.84 & 0.79 & 576 \\ \cline{2-6}
                                                & 2 & 0.86 & 0.81 & 0.84 & 625 \\ \cline{2-6}
                                                & 3 & 0.82 & 0.74 & 0.78 & 416 \\ \cline{2-6}
                                                & 4 & 0.79 & 0.79 & 0.79 & 578 \\ 
        \hline
        \multirow{4}{*}{\textbf{CatBoost}} & 1 & 0.77 & 0.84 & 0.80 & 576 \\ \cline{2-6}
                                             & 2 & 0.85 & 0.83 & 0.84 & 625 \\ \cline{2-6}
                                             & 3 & 0.85 & 0.77 & 0.81 & 416 \\ \cline{2-6}
                                             & 4 & 0.79 & 0.80 & 0.79 & 578 \\ 
        \hline
        \multicolumn{6}{|c|}{\textbf{Testing Scores}} \\ 
        \hline
        \multirow{4}{*}{\textbf{Proposed Model}} & 1 & 0.77 & 0.79 & 0.78 & 321 \\ \cline{2-6}
                                                & 2 & 0.87 & 0.83 & 0.85 & 294 \\ \cline{2-6}
                                                & 3 & 0.59 & 0.78 & 0.67 & 232 \\ \cline{2-6}
                                                & 4 & 0.82 & 0.65 & 0.73 & 352 \\ 
        \hline
        \multirow{4}{*}{\textbf{Random Forest}} & 1 & 0.65 & 0.78 & 0.71 & 321 \\ \cline{2-6}
                                                & 2 & 0.87 & 0.71 & 0.78 & 294 \\ \cline{2-6}
                                                & 3 & 0.54 & 0.77 & 0.64 & 232 \\ \cline{2-6}
                                                & 4 & 0.76 & 0.55 & 0.64 & 352 \\ 
        \hline
        \multirow{4}{*}{\textbf{CatBoost}} & 1 & 0.74 & 0.81 & 0.77 & 321 \\ \cline{2-6}
                                             & 2 & 0.88 & 0.79 & 0.83 & 294 \\ \cline{2-6}
                                             & 3 & 0.59 & 0.80 & 0.68 & 232 \\ \cline{2-6}
                                             & 4 & 0.80 & 0.60 & 0.68 & 352 \\ 
        \hline
    \end{tabular}
\end{table}

\section{Limitations and Future Scopes}
In this work, we make several key contributions to advance machine learning applications. First, we prioritize traditional machine learning (ML) techniques over deep learning (DL), enabling reduced memory and data requirements for training and testing, which leads to significantly shorter training times. We introduce a customized dataset tailored to our specific application and propose a novel ML model that effectively utilizes this data. Additionally, we incorporate a camera correlation factor to enhance accuracy across different perspectives and leverage multi-view data integration to create a coherent representation in non-overlapping situations. Our model is designed to perform robustly in complex backgrounds and varying lighting conditions, addressing challenges often overlooked in existing datasets. Through these contributions, we aim to provide an efficient and effective solution for real-world scenarios. Our gait-based person identification and re-identification model provides a robust baseline for real-life dataset analysis, offering advantages over traditional appearance-based methods by being effective in varied scenarios such as video surveillance and human-robot interaction. Despite its promise, the model faces limitations in detecting individuals engaged in dynamic movements like running or jumping and struggles with accuracy in crowded environments where gait patterns are less distinct. Future developments could include creating a real-time system that integrates video input with preprocessing and classification, enhancing multiple person detection capabilities, and combining various recognition methods to improve accuracy and handle irregular data.

\section{Conclusion}
Our gait-based person re-identification model, utilizing non-overlapping surveillance cameras, addresses real-life challenges with a focus on cost-efficiency and robustness in uncontrolled environments. By analyzing Light Gradient Boosting Machine (LGBM), Random Forest, and CatBoost, we identified LGBM as particularly effective for our scenario. This work sets the stage for future advancements, including enhanced datasets, ethical considerations, and sophisticated deep learning techniques. The future holds promise for integrating 3D gait analysis and multiple biometric modalities, potentially revolutionizing gait re-identification systems in computer vision.

\end{document}